# Convex Structure Learning for Bayesian Networks: Polynomial Feature Selection and Approximate Ordering

**Yuhong Guo** and **Dale Schuurmans**
Department of Computing Science
University of Alberta
{yuhong,dale}@cs.ualberta.ca

## Abstract

We present a new approach to learning the structure and parameters of a Bayesian network based on regularized estimation in an exponential family representation. Here we show that, given a fixed variable order, the optimal structure and parameters can be learned efficiently, even without restricting the size of the parent variable sets. We then consider the problem of optimizing the variable order for a given set of features. This is still a computationally hard problem, but we present a convex relaxation that yields an optimal "soft" ordering in polynomial time. One novel aspect of the approach is that we do not perform a discrete search over DAG structures, nor over variable orders, but instead solve a continuous convex relaxation that can then be rounded to obtain a valid network structure. We conduct an experimental comparison against standard structure search procedures over standard objectives, which cope with local minima, and evaluate the advantages of using convex relaxations that reduce the effects of local minima.

## 1 Introduction

Bayesian networks are one of the most prevalent and useful formalisms for representing uncertain knowledge (Pearl, 1988). Along with Markov networks, they share the advantage of providing a sound probabilistic foundation for inference and learning, and can represent complex distributions compactly. However, Bayesian networks offer a distinct advantage in interpretability, since each parameter can be interpreted in isolation as a conditional probability assertion over a subset of variables in the domain. They also offer computational benefits over Markov networks, by permitting more efficient parameter estimation for example.

Learning a Bayesian network from data poses the problem of estimating the parameters of the model, and more interestingly, to infer the structure of the network. Parameter estimation from complete data is a generally well understood problem that permits effective algorithmic approaches. Structure learning, on the other hand, is a much more challenging problem that has yet to yield a completely satisfactory solution. A key issue in learning structure is to develop a reasonable principle for model selection, since a complex network structure is always able to fit training data better than a simpler structure. Beyond statistical questions, however, structure learning poses a significantly harder computational problem, since one must cope with a combinatorial search over the space of possible structures.

The two main approaches taken to Bayesian network structure learning are commonly referred to as constraint based and score based respectively. In the constraint based approach, one first attempts to identify a set of conditional independence properties that hold in the domain, and then attempts to identify a network structure that best satisfies these constraints (Spirtes et al., 2000). The difficulty with this approach is that reliably identifying the conditional independence properties, and optimizing the network structure are both challenging problems (Margaritis, 2003). A much more common approach is the score based method, where one first posits a criterion by which a given Bayesian network structure can be evaluated on a given data set. The task is then to search for a Bayesian network structure that optimizes the score. Since model selection is such a critical issue, score based approaches are typically based on well established statistical principles such as minimum description length (MDL) (Rissanen, 1987; Lam & Bacchus, 1994; Friedman & Goldszmidt, 1998; Van Allen & Greiner, 2000) or Bayesian estimation. The use of Bayesian scoring approaches was developed in (Cooper & Herskovits, 1992), culminating in the BDe score of (Heckerman et al., 1995), which is currently one of the best known standards. These scores offer sound, well motivated model selection criteria for Bayesian network structure. The main problem with using these scores, however, is that they create intractable optimization problems. That is, it is NP-hard to compute the optimal network for the Bayesian scores (Chickering, 1996). In fact, recently it has been shown that optimizing Bayesian network structure is NP-hard, in the large sample limit, for all consistent scoring criteria, including MDL, BIC and the Bayesian scores (Chickering et al., 2003).

Due to the known intractability of structure optimization, the literature on Bayesian network structure learning has been dominated by heuristic algorithms for searching the space of networks, including greedy local search, random re-starts, simulated annealing and genetic algorithms (Goldenberg & Moore, 2004; Moore & Wong, 2003; Heckerman et al., 1995; Elidan et al., 2002; Larranaga et al., 1996). However, recently it has been observed that searching the space of variable orderings can be more effective than searching the space of network structures (Larranaga et al., 1996; Teyssier & Koller, 2005), since the space of orderings is much smaller. This approach exploits the fundamental insight of (Cooper & Herskovits, 1992; Buntine, 1991) that, for a fixed variable order, the optimal network (of bounded in-degree) and parameters can be computed in polynomial time (but exponential in the in-degree bound).

In this paper we offer an alternative approach to the problem of learning a Bayesian network model from data. Our idea is to follow the strategy from combinatorial optimization, where, when faced with an intractable combinatorial problem, one first formulates a convex relaxation that can be solved efficiently, and then rounds the "soft" solution to obtain an approximate "hard" solution to the original problem. Here, we propose an efficient relaxation of the Bayesian network structure learning problem—solving for the structural features that determine the graph, the variable ordering that determines the edge orientation, and the model parameters—in a single, compact optimization. First, we show that, given a fixed variable order, the maximum likelihood structure and parameters can be found in polynomial time and space using a sparse exponential family representation, without any restriction on the number of parents for any variable. Second, given a fixed variable order, we show how feature selection based on the minimum description length principle can be addressed simultaneously with parameter optimization. Finally, to optimize the order, we introduce a compact matrix representation of total orderings that allows a tight semidefinite relaxation. We evaluate our overall technique on natural and synthetic data sets, and find that convex relaxation is a very promising approach to this problem, even though the underlying search problem is inherently discrete.

## 2 Bayesian network representations

A Bayesian network is normally defined by a directed acyclic graph over variables $X_1, ..., X_n$, where the probability of a configuration $\mathbf{x}$ is given by

$$\begin{aligned} P(\mathbf{x}) &= \prod_{j=1}^{n} P(x_j | \mathbf{x}_{\pi(j)}) \\ &= \exp\Big( \sum_{ja\mathbf{b}} 1_{(\mathbf{x}_j = a\mathbf{b})} \ln \theta_{ja\mathbf{b}} \Big) \end{aligned} \quad (1)$$

Here $\boldsymbol{\theta}$ denotes the parameters of the model; $j$ ranges over conditional probability tables (CPTs), one for each variable $X_j$; $1_{(\cdot)}$ denotes the indicator function; $\mathbf{x}_j$ denotes the local subconfiguration of $\mathbf{x}$ on $(x_j, \mathbf{x}_{\pi(j)})$; $a$ denotes the set of values for child variable $x_j$; and $\mathbf{b}$ denotes the set of configurations for $x_j$'s parents $\mathbf{x}_{\pi(j)}$. The form (1) shows how Bayesian networks are a form of exponential model $P(\mathbf{x}) = \exp\left(\mathbf{w}^\top \boldsymbol{\phi}(\mathbf{x})\right)$, using the substitution $w_{ja\mathbf{b}} = \ln \theta_{ja\mathbf{b}}$. Here $\boldsymbol{\phi}(\mathbf{x})$ denotes the feature vector $(...1_{(\mathbf{x}_j = a\mathbf{b})}...)^\top$ over $j, a, \mathbf{b}$. In fact, we will adopt a general exponential form representation for Bayesian networks in this paper, since we will exploit many of the advantages it offers over the traditional CPT based representation.

Rather than start with CPT entries $\boldsymbol{\theta}$, one can alternatively represent a Bayesian network in a general exponential form

$$P(\mathbf{x}) = \exp\left( \sum_j \left[ \mathbf{w}_j^\top \boldsymbol{\phi}_j(x_j, \mathbf{x}_{\pi(j)}) - A(\mathbf{w}_j, \mathbf{x}_{\pi(j)}) \right] \right)$$

where

$$A(\mathbf{w}_j, \mathbf{x}_{\pi(j)}) = \log\left( \sum_a \exp\left(\mathbf{w}_j^\top \boldsymbol{\phi}_j(a, \mathbf{x}_{\pi(j)})\right) \right)$$

Here $A(\mathbf{w}_j, \mathbf{x}_{\pi(j)})$ is the log normalization constant for the $j$th conditional probability distribution. We use $\boldsymbol{\phi}_j(x_j, \mathbf{x}_{\pi(j)})$ to denote the feature vector $(...1_{(x_j = a, \mathbf{x}_{\pi(j)} = \mathbf{b})}...)^\top$ over $a, \mathbf{b}$, and use $\mathbf{w}_j$ to denote the local weight vector $(...w_{ja\mathbf{b}}...)^\top$ over $a, \mathbf{b}$. Thus, together $\boldsymbol{\phi}_j$ and $\mathbf{w}_j$ specify the local conditional probability distribution $P(x_j | \mathbf{x}_{\pi(j)})$ and allow the traditional CPT parameter entries to be efficiently recovered by $\theta_{ja\mathbf{b}} = \exp(\mathbf{w}_j^\top \boldsymbol{\phi}_j(a, \mathbf{b}) - A(\mathbf{w}_j, \mathbf{b}))$.

As we will see below, one key aspect of the exponential form is that it expresses $P(\mathbf{x})$ as a *convex* function of the parameters $\mathbf{w}$, which will lead to convenient optimization problems. Another important advantage of the exponential form, however, is that it allows a *sparse* representation of the conditional distributions. That is, we can represent $P(x_j | \mathbf{x}_{\pi(j)})$ given a subset of features from the set of possibilities $\{1_{(x_j = a, \mathbf{x}_{\pi(j)} = \mathbf{b})} : a \in \textit{Vals}(x_j), \mathbf{b} \in \textit{Vals}(\mathbf{x}_{\pi(j)})\}$. In general, this allows one to represent $P(x_j | \mathbf{x}_{\pi(j)})$ compactly even if the number of parent variables is large. Such a sparse feature representation of a CPT is similar to exploiting context specific independence (Boutilier et al., 1996) or local structure (Friedman & Goldszmidt, 1998). In fact, these compact representations can be recovered as a special case. The size of a feature based representation for a CPT is never larger than the traditional table based representation, and can be arbitrarily smaller.

Below we find that the feature based representation is particularly advantageous from the perspective of learning a Bayesian network from data, since it nicely reduces the problem of structure learning, largely, to identifying the features used to define the model. The only other issue is to acquire an effective ordering for the variables in the model. Overall, the problem of learning a Bayesian network from data can be decomposed into the three problems of: learning a set of features, learning a variable ordering, and learning a corresponding set of parameters. Below we propose an approach that tackles all three subproblems simultaneously.

## 3 Parameter estimation

Before tackling each subproblem in turn, we first establish some preliminary results that will be needed later. The first and simplest subproblem is estimating the parameters $\mathbf{w}$ given a fixed ordering $\pi$ and feature set $\phi$.

Given complete training data $D = [\mathbf{x}^1; ...; \mathbf{x}^T]$, the negative loglikelihood loss can be expressed

$$L(\mathbf{w}) = \sum_{i,j} \left[ A(\mathbf{w}_j, \mathbf{x}^i_{\pi(j)}) - \mathbf{w}_j^\top \phi_j(x^i_j, \mathbf{x}^i_{\pi(j)}) \right]$$

$$= \sum_{j, \mathbf{b}_j} \#\mathbf{b}_j \left[ A(\mathbf{w}_j, \mathbf{b}_j) - \mathbf{w}_j^\top \bar{\phi}_{\mathbf{b}_j} \right]$$

where $\bar{\phi}_{\mathbf{b}_j} = \sum_{a_j} \frac{\#(a_j \mathbf{b}_j)}{\#\mathbf{b}_j} \phi_j(a_j, \mathbf{b}_j)$. Since $A(\mathbf{w}_j, \mathbf{b})$ is a convex function of $\mathbf{w}_j$, this leads to a convex minimization problem for $\mathbf{w}_j$. However, since overfitting is always a concern, we will find it advantageous to minimize the regularized loglikelihood loss

$$\tilde{L}(\mathbf{w}) = \frac{\beta}{2}\|\mathbf{w}\|^2 + L(\mathbf{w}) = \qquad (2)$$

$$\sum_j \left( \frac{\beta}{2}\|\mathbf{w}_j\|^2 + \sum_{\mathbf{b}_j} \#\mathbf{b}_j \left[ A(\mathbf{w}_j, \mathbf{b}_j) - \bar{\phi}_{\mathbf{b}_j}^\top \mathbf{w}_j \right] \right)$$

Here $\beta$ is a regularization parameter. Note that the weights that minimize $\tilde{L}(\mathbf{w})$ correspond to a MAP estimate of $\mathbf{w}$, with prior $\mathbf{w} \sim \mathcal{N}(\mathbf{0}, \beta I)$.

The objective (2) decomposes as an independent sum over $j$, so we can consider the minimization of each individual objective separately. To reduce the notational burden, denote the $j$th component of $\tilde{L}(\mathbf{w})$ by

$$\tilde{L}(\mathbf{u}) = \frac{\beta}{2}\|\mathbf{u}\|^2 + \sum_{\mathbf{b}} \#\mathbf{b} \left[ A(\mathbf{u}, \mathbf{b}) - \bar{\phi}_{\mathbf{b}}^\top \mathbf{u} \right] \quad (3)$$

Although $\tilde{L}(\mathbf{u})$ is a convex minimization objective, it turns out that to derive our results below we will need to work with the *dual*. The dual is derived by formulating a tight concave lower bound on $\tilde{L}(\mathbf{u})$, which can then be maximized to recover an equivalent result to $\tilde{L}(\mathbf{u})$. A tight lower bound can easily be derived in this case using the convex conjugate function for $A(\mathbf{u}, \mathbf{b})$, given by

$$A^*(\boldsymbol{\mu}, \mathbf{b}) = \max_{\mathbf{u}} \boldsymbol{\mu}_{\mathbf{b}}^\top \mathbf{u} - A(\mathbf{u}, \mathbf{b}) = \sum_a \mu_{a\mathbf{b}} \log \mu_{a\mathbf{b}}$$

$$= -H(\boldsymbol{\mu}_{\mathbf{b}}) \quad \text{where } \boldsymbol{\mu}_{\mathbf{b}} \geq 0 \text{ and } \mathbf{e}^\top \boldsymbol{\mu}_{\mathbf{b}} = 1$$

Here we use $\mathbf{e}$ to denote the vector of all 1s. This function is also convex (Boyd & Vandenberghe, 2004), and the dual variables $\boldsymbol{\mu}$ satisfy the relation $\mu_{a\mathbf{b}} = E[\phi_{a\mathbf{b}}(x)|\mathbf{b}]$ (Wainwright & Jordan, 2003).

Since our Bayesian network representation is based on using *indicator* features, $\phi_{a\mathbf{b}}(x, \mathbf{x}_\pi) = 1_{(x=a, \mathbf{x}_\pi = \mathbf{b})}$, we can furthermore obtain a convenient one-to-one relationship between the dual variables $\mu_{a\mathbf{b}}$ and the CPT parameter entries $\theta_{a\mathbf{b}}$ by noting that $\mu_{a\mathbf{b}} = E[\phi_{a\mathbf{b}}(x)|\mathbf{x}_\pi = \mathbf{b}] = P(x=a|\mathbf{x}_\pi = \mathbf{b}) = \theta_{a\mathbf{b}}$. Therefore, we can think of the convex conjugate function for $A(\mathbf{u}, \mathbf{b})$ as being expressed in the CPT parameters directly:

$$A^*(\boldsymbol{\theta}, \mathbf{b}) = -H(\boldsymbol{\theta}_{\mathbf{b}}) \quad \text{where } \boldsymbol{\theta}_{\mathbf{b}} \geq 0 \text{ and } \mathbf{e}^\top \boldsymbol{\theta}_{\mathbf{b}} = 1$$

The key property that the conjugate function provides is that it establishes a *concave* lower bound. In fact, since $A(\mathbf{u}, \mathbf{b})$ is convex in $\mathbf{u}$ it can be shown that

$$A(\mathbf{u}, \mathbf{b}) = \max_{\boldsymbol{\theta}_{\mathbf{b}} \geq 0, \, \mathbf{e}^\top \boldsymbol{\theta}_{\mathbf{b}} = 1} \boldsymbol{\theta}_{\mathbf{b}}^\top \mathbf{u} + H(\boldsymbol{\theta}_{\mathbf{b}})$$

(Wainwright & Jordan, 2003). Using this fact we obtain

$$\min_{\mathbf{u}} \tilde{L}(\mathbf{u}) \qquad (4)$$

$$= \min_{\mathbf{u}} \frac{\beta}{2}\|\mathbf{u}\|^2 + \sum_{\mathbf{b}} \#\mathbf{b} \left[ \max_{\boldsymbol{\theta}_{\mathbf{b}}} \boldsymbol{\theta}_{\mathbf{b}}^\top \mathbf{u} + H(\boldsymbol{\theta}_{\mathbf{b}}) - \mathbf{u}^\top \bar{\phi}_{\mathbf{b}} \right]$$

$$= \max_{\boldsymbol{\theta}} \min_{\mathbf{u}} \frac{\beta}{2}\|\mathbf{u}\|^2 + \sum_{\mathbf{b}} \#\mathbf{b} \left[ H(\boldsymbol{\theta}_{\mathbf{b}}) - \mathbf{u}^\top (\bar{\phi}_{\mathbf{b}} - \boldsymbol{\theta}_{\mathbf{b}}) \right]$$

Note that the minimum and the maximum can be interchanged here because the problem is convex and there is no duality gap (Boyd & Vandenberghe, 2004). Taking the derivative of the inner objective with respect to $\mathbf{u}$ yields

$$\nabla_{\mathbf{u}} = \beta \mathbf{u} - \sum_{\mathbf{b}} \#\mathbf{b} (\bar{\phi}_{\mathbf{b}} - \boldsymbol{\theta}_{\mathbf{b}}) = 0, \text{ so that}$$

$$\mathbf{u}^*(\boldsymbol{\theta}) = \frac{1}{\beta} \sum_{\mathbf{b}} \#\mathbf{b} (\bar{\phi}_{\mathbf{b}} - \boldsymbol{\theta}_{\mathbf{b}}), \text{ and therefore} \qquad (5)$$

$$\tilde{L}(\boldsymbol{\theta}) = \qquad (6)$$

$$\max_{\boldsymbol{\theta}} \left[ \sum_{\mathbf{b}} \#\mathbf{b} H(\boldsymbol{\theta}_{\mathbf{b}}) \right] - \frac{1}{2\beta} \left\| \sum_{\mathbf{b}} \#\mathbf{b} (\bar{\phi}_{\mathbf{b}} - \boldsymbol{\theta}_{\mathbf{b}}) \right\|^2$$

where $\boldsymbol{\theta} \geq 0$, $\mathbf{e}^\top \boldsymbol{\theta}_{\mathbf{b}} = 1$ for all $\mathbf{b}$. Thus the dual to the minimum regularized loglikelihood loss problem is a regu-larized concave maximum entropy problem. Given a solution to the dual problem $\theta^*$ a corresponding primal solution can be easily recovered using (5).

For implementation, the primal problem is more convenient than the dual because it is unconstrained. In our implementation below we used a Newton method to efficiently solve (3). The dual formulation is required to establish our theoretical results below, however.[1]

## 4 Strategy

Of course, the main goal in this paper is not to perform parameter estimation, but to learn the structure of a Bayesian network model from data. The exponential family representation and maximum entropy frameworks offer a new perspective on this problem. Rather than scoring a Bayesian network and performing a discrete search in structure space, our goal will be to formulate a polynomial time approach that addresses each of the three subproblems—feature generation (and selection), parameter estimation, and variable ordering—in a joint convex optimization that relies on reasonable convex relaxations of the discrete subproblems when necessary.

We pursue the following strategy. First, we generate a "universal" set of features that allows us to express any maximum likelihood solution exactly. Our first result below shows that in fact this can be achieved in polynomial time and space given a fixed variable ordering. Second, we select a subset of the generated features using the minimum description length principle (Rissanen, 1987; Lam & Bacchus, 1994). Our main result here is that, using the maximum entropy estimation framework developed above, MDL feature selection and parameter optimization can be performed simultaneously in a novel convex relaxation. Finally, we include variable ordering in the framework by extending the previous optimization formulation to also search over variable orders. Our third main result is that a search over variable orders can be efficiently encoded by a compact set of semidefinite constraints on a matrix representation of the ordering. Overall, we are able to solve a relaxed form of the entire Bayesian network learning problem within a polynomial convex optimization framework.

## 5 Feature generation

Our first result is that, given a fixed variable order, a set of features sufficient to represent any maximum likelihood Bayesian network can be found in polynomial time and is polynomially large. This result holds without restriction on the number of parents of any variable. In fact, the result is straightforward, but relies heavily on the sparse feature representation. They key idea is that one can use linear dependence of feature responses on augmented training data to identify key features and eliminate other features from consideration.

First, note that since the conditional probabilities are locally defined and the variable ordering is known, we can solve the feature generation problem for each variable $x_j$ independently. Next, assume that the variable indices are sorted according to the ordering so that the set of possible parents of $x_j$ is $\{x_1, ..., x_{j-1}\}$. Let $\sigma(j) = \{1, ..., j-1\}$ denote the set of ancestors of $j$ under the ordering. Then given a set of complete training data (row vectors) represented in a $T \times n$ data matrix, $D = [\mathbf{x}^1; ...; \mathbf{x}^T]$, only the first $j$ columns of $D$ are relevant for $x_j$.

To identify a universal set of features, it suffices to consider a locally augmented data matrix where we copy each ancestor configuration, $\mathbf{x}^i_{\sigma(j)}$, $V_j$ times and replace $x^i_j$ with each of its possible values. Here $V_j = |\textit{Vals}(x_j)|$. Call the resulting matrix $\tilde{D}_j$; so if $D_j$ is $T \times j$ then $\tilde{D}_j$ is $(TV_j) \times j$.

**Proposition 1** *For any two exponential form representations $\phi_1, \mathbf{w}_1$ and $\phi_2, \mathbf{w}_2$: if $\mathbf{w}_1^\top \phi_1(a, \mathbf{x}^i_{\sigma(j)}) = \mathbf{w}_2^\top \phi_2(a, \mathbf{x}^i_{\sigma(j)})$ for all $i = 1, ..., T$ and all $a \in \textit{Vals}(x_j)$, then $P_1(x^i_j | \mathbf{x}^i_{\sigma(j)}) = P_2(x^i_j | \mathbf{x}^i_{\sigma(j)})$ for all $i = 1, ..., T$.*

**Proof** First note that the assumption implies that $A_1(\mathbf{w}_1, \mathbf{x}^i_{\sigma(j)}) = \log \sum_a \exp\left(\mathbf{w}_1^\top \phi_1(a, \mathbf{x}^i_{\sigma(j)})\right) = \log \sum_a \exp\left(\mathbf{w}_2^\top \phi_2(a, \mathbf{x}^i_{\sigma(j)})\right) = A_2(\mathbf{w}_2, \mathbf{x}^i_{\sigma(j)})$ for all $i$. Therefore we must also have $-\log P_1(x^i_j | \mathbf{x}^i_{\sigma(j)}) = A_1(\mathbf{w}_1, \mathbf{x}^i_{\sigma(j)}) - \mathbf{w}_1^\top \phi_1(a, \mathbf{x}^i_{\sigma(j)}) = A_2(\mathbf{w}_2, \mathbf{x}^i_{\sigma(j)}) - \mathbf{w}_2^\top \phi_2(a, \mathbf{x}^i_{\sigma(j)}) = -\log P_2(x^i_j | \mathbf{x}^i_{\sigma(j)})$. ∎

Thus if one set of features $\phi_1$ spans another set $\phi_2$ on the augmented data matrix $\tilde{D}_j$ for each variable $x_j$, then the optimal parameter estimate for $\phi_1$ (either maximum likelihood or regularized likelihood) on $D$ has to be at least as good as the best parameter estimate for $\phi_2$; that is, $\tilde{L}(\mathbf{w}_1^*, \phi_1, D) \leq \tilde{L}(\mathbf{w}_2^*, \phi_2, D)$.

Of course, there are many possible features to consider. There is a unique feature $\phi_{ja\mathbf{b}}$ corresponding to an indicator function $\phi_{ja\mathbf{b}}(x_j, \mathbf{x}_{\rho(j)}) = 1_{(x_j = a, \mathbf{x}_{\rho(j)} = \mathbf{b})}$ for each particular subset of ancestor variables, $\rho(j) \subset \sigma(j)$, and each particular value $a$ for $x_j$ and value $\mathbf{b}$ for $\mathbf{x}_{\rho(j)}$. Nevertheless, it is a trivial observation that the maximum rank of any possible span is bounded by $TV_j$, since this is the length of each feature response vector on the augmented training set $\tilde{D}_j$. Therefore, there must exist a set of no more than $TV_j$ features that allows the exponential form representation achieve the maximum likelihood score (or regularized likelihood score) of *any* Bayesian network on the training data $D_j$.

---

[1] We note that in the Bayesian network learning literature it is common to adopt a Bayesian perspective on parameter estimation as well as structure learning (Heckerman et al., 1995; Cooper & Herskovits, 1992). For parameter estimation, the standard approach is to use a Dirichlet prior over each of the conditional probability vectors $\theta_{j\mathbf{b}}$, with corresponding prior parameters $\alpha_{j\mathbf{b}}$. Here the posterior mean estimate is given by $\hat{\theta}_{ja\mathbf{b}} = (\#(ja\mathbf{b}) + \alpha_{ja\mathbf{b}})/(\#(j\mathbf{b}) + \sum_a \alpha_{ja\mathbf{b}})$. Although the Bayesian posterior estimate appears to be quite different from the solution to (6), the two estimates in fact behave similarly. For example, in the large sample limit both estimates $\hat{\theta}_{j\mathbf{b}}$ converge to $E[\phi_j | \mathbf{b}]$. With no data and uniform $\alpha$, both approaches produce the same maximum entropy estimates. The advantage of the quadratic regularizer in (6) is that it allows us to express a convex formulation of the minimum description length principle for structure learning, as we will see below.

To find this set of features in polynomial time we exploit the fact that every compound feature $\phi_{ja\mathbf{b}}$ can be decomposed as a product for features defined on shorter patterns

$$
\begin{aligned}
\phi_{ja\mathbf{b}}(x_j, \mathbf{x}_{\rho(j)}) \\
&= 1_{(x_j=a, \mathbf{x}_{\rho(j)}=\mathbf{b})} \quad (7) \\
&= 1_{(x_j=a)} 1_{(x_{\rho_1(j)}=b_1)} \ldots 1_{(x_{\rho_k(j)}=b_k)} \\
&= \phi_{ja}(x_j) \phi_{\rho_1(j)b_1}(x_{\rho_1(j)}) \ldots \phi_{\rho_k(j)b_k}(x_{\rho_k(j)})
\end{aligned}
$$

Naturally we would like to build a span consisting of the shortest possible feature patterns, since this would result in a simpler Bayesian network representation. Define the length of $\phi_{ja\mathbf{b}}$ to be the number of variables in its definition (7). Then we have the following proposition.

**Proposition 2** *If a compound feature $\phi_{ja\mathbf{b}}$ is spanned by a set of shorter features, then $\phi_{ja\mathbf{b}}$ is unnecessary.*

**Proof** Assume $\phi_{ja\mathbf{b}} = \sum_f w_f \phi_f$ on $\tilde{D}_j$ for some set of shorter feature patterns $f \in F$. Then any extended feature that uses $\phi_{ja\mathbf{b}}$ can be spanned by features based on shorter patterns. In particular, if $\phi_{jc\mathbf{d}} = \phi_{ja\mathbf{b}} \phi_{g_1} \ldots \phi_{g_k}$ on $\tilde{D}_j$, then we must also have $\phi_{jc\mathbf{d}} = \sum_f w_f \phi_f \phi_{g_1} \ldots \phi_{g_k}$ on $\tilde{D}_j$, where the feature patterns in the second expansion are strictly shorter than the first. ∎

This leads to a polynomial time algorithm for generating a set of shortest features with maximum span on $\tilde{D}_j$; see Figure 1. To establish that this procedure does indeed run in polynomial time, consider the lattice of feature patterns. The lattice is searched from shortest patterns to longest. Once a pattern is pruned, no extension of it will ever be considered (and correctness will be preserved by Proposition 2). However, for each increase in rank, at most $\sum_{\ell=1}^{j} \textit{Vals}(x_\ell)$ features are added, while the maximum rank is $TV_j$.

One drawback of this procedure is that it can generate a large number of parents for $x_j$, even though the representation remains polynomially large. In fact, this feature generation process is guaranteed to overfit the data, in the sense that it yields a representation that can achieve the maximum likelihood of *any* Bayesian network. Clearly, some sort of feature selection process is required to yield a reasonable model, which we now consider.

*Feature generation procedure for $x_j$ on augmented $\tilde{D}_j$:*
$\Phi^{(0)} = \{\phi_\emptyset\}$ (the constant 1 feature)
for $k = 1, 2, \ldots$ (while rank increased)
    $\Phi^{(k)} \leftarrow \emptyset$
    for each $\phi_f \in \Phi^{(k-1)}$
        $\Psi \leftarrow \{\phi_{b_\ell} \phi_f : \ell \notin f, b_\ell \in \textit{Vals}(x_\ell)\}$
        if rank $\left(\bigcup_{\ell \leq k} \Phi^{(\ell)} \cup \Psi\right) >$ rank $\left(\bigcup_{\ell \leq k} \Phi^{(\ell)}\right)$
            $\Phi^{(k)} \leftarrow \Phi^{(k)} \cup \Psi$

Figure 1: Feature generation procedure

## 6 Feature selection

We base our feature selection strategy on the minimum description length principle (Rissanen, 1987; Lam & Bacchus, 1994; Friedman & Goldszmidt, 1998). Here we continue to assume a fixed variable ordering is given. The idea is to start with a large set of "universal" features $\boldsymbol{\phi} = (\ldots \phi_{ja\mathbf{b}} \ldots)^\top$ generated by the procedure outlined previously. To perform feature selection in this large set, we augment the exponential representation with feature selection variables $\boldsymbol{\eta}$. That is, for each feature $\phi_{ja\mathbf{b}}$ we establish a corresponding selector variable $\eta_{ja\mathbf{b}} \in \{0, 1\}$, in addition to the corresponding weight $w_{ja\mathbf{b}}$. Let $N_j = \text{diag}(\boldsymbol{\eta}_j)$ be the diagonal matrix of selector values corresponding to variable $x_j$. We can then write

$$
P(\mathbf{x}) = \exp\left(\sum_j \left[\mathbf{w}_j^\top N_j \phi_j(x_j, \mathbf{x}_{\pi(j)}) - A(\mathbf{w}_j, \mathbf{x}_{\pi(j)})\right]\right)
$$

$$
A(\mathbf{w}_j, \mathbf{x}_{\pi(j)}) = \log \sum_a \exp\left(\mathbf{w}_j^\top N_j \phi_j(a, \mathbf{x}_{\pi(j)})\right)
$$

If $\eta_{ja\mathbf{b}} = 1$ then the feature $\phi_{ja\mathbf{b}}$ is selected, otherwise it is dropped.

We would like to solve for the set of features $\boldsymbol{\eta}$ and parameters $\mathbf{w}$ that minimize the total description length of the data and the Bayesian network model (in exponential form). This can be formulated as a joint optimization of

$$
\min_{\boldsymbol{\eta} \in \{0,1\}^F} \mathbf{c}^\top \boldsymbol{\eta} + \frac{\log T}{2} \mathbf{e}^\top \boldsymbol{\eta} + \min_{\mathbf{w}} \tilde{L}(\mathbf{w}, \boldsymbol{\eta}, D) \quad (8)
$$

Here the last term is the cost of encoding the training data $D = [\mathbf{x}^1; \ldots; \mathbf{x}^T]$ using the optimal parameters $\mathbf{w}$ for the network structure specified by $\boldsymbol{\eta}$. This uses a standard result from information theory (Cover & Thomas, 1991; Friedman & Goldszmidt, 1998) that an optimal code for data $D$ given a model $P(\mathbf{x})$ has length $-\sum_i \log P(\mathbf{x}^i)$. (Although here we alter this principle slightly to use the regularlized loss $\tilde{L}$ rather than the plain loglikelihood loss $L$. This simplifies the derivation below.)

The first term in (8) measures the length of the description for an exponential family representation for the Bayesian network structure specified by $\boldsymbol{\eta}$. In particular, for each feature $\phi_{ja\mathbf{b}}$ selected by indicator $\eta_{ja\mathbf{b}}$ we fix the description length cost

$$c_{ja\mathbf{b}} = |\mathbf{b}|\log n + \log|\mathit{Vals}(a)| + \sum_\ell \log|\mathit{Vals}(b_\ell)|$$

where the first term is the cost of encoding the list of variables in feature $\phi_{ja\mathbf{b}}$, and the remaining terms are the cost of encoding the specific values for these variables.

The second term in (8) is the cost of encoding each weight parameter $w_{ja\mathbf{b}}$, where the precision is chosen in the manner discussed in (Friedman & Goldszmidt, 1998).

Now we would like to solve for the structure $\boldsymbol{\eta}$ and parameters $\mathbf{w}$ that minimize the total description cost (8). Unfortunately, this is a combinatorial optimization problem over $\boldsymbol{\eta}$, and even more problematic, even if $\boldsymbol{\eta}$ were relaxed, the MDL objective (8) is not jointly convex in $\boldsymbol{\eta}$ and $\mathbf{w}$. Fortunately, the *dual* form of the regularized loss allows us to re-express this problem as a convex minimization over $\boldsymbol{\eta}$, ignoring the integer constraints.

Using the fact that (6) is equivalent to (4), yields an equivalent optimization problem to (8), but now using maximum entropy instead of loglikelihood loss:

$$\min_{\boldsymbol{\eta}\in\{0,1\}^F} \mathbf{c}^\top\boldsymbol{\eta} + \frac{\log T}{2}\mathbf{e}^\top\boldsymbol{\eta} + \quad (9)$$

$$\max_{\boldsymbol{\theta}} \sum_j \left(\left[\sum_\mathbf{b} \#\mathbf{b} H(\boldsymbol{\theta}_\mathbf{b})\right] - \frac{1}{2\beta}\boldsymbol{\delta}_j^\top N_j^\top N_j \boldsymbol{\delta}_j\right)$$

where $\boldsymbol{\delta}_j = \sum_{\mathbf{b}_j} \#\mathbf{b}_j(\bar{\boldsymbol{\phi}}_{\mathbf{b}_j} - \boldsymbol{\theta}_{\mathbf{b}_j})$. Recall that, thus far, we have assumed that $\boldsymbol{\eta} \in \{0,1\}^F$, and therefore $N_j^\top N_j = N_j$, since $n_{ja\mathbf{b}}^2 = \eta_{ja\mathbf{b}}$. This allows us to rewrite (9) as $\min_{\boldsymbol{\eta}} g(\boldsymbol{\eta})$ where

$$g(\boldsymbol{\eta}) = \max_{\boldsymbol{\theta}} \mathbf{c}^\top\boldsymbol{\eta} + \frac{\log T}{2}\mathbf{e}^\top\boldsymbol{\eta} + \quad (10)$$

$$\sum_j \left(\left[\sum_\mathbf{b} \#\mathbf{b} H(\boldsymbol{\theta}_\mathbf{b})\right] - \frac{1}{2\beta}\boldsymbol{\delta}_j^\top N_j \boldsymbol{\delta}_j\right)$$

Crucially, $g(\boldsymbol{\eta})$ is a pointwise maximum of *linear* functions of $\boldsymbol{\eta}$, and is therefore convex (Boyd & Vandenberghe, 2004).

Thus, by combining regularized maximum entropy parameter estimation with the description length penalty, we obtain a natural convex relaxation of the minimum description length principle simply by relaxing the structure indicator variables to be soft indicators in the interval $[0,1]$. Remarkably, this formulation allows one to simultaneously optimize the (soft) structure and parameters in a polynomial size convex optimization problem.

To solve this problem in practice, we use a quasi-Newton method, BFGS (Nocedal & Wright, 1999) with backtracking line search to efficiently minimize $g(\boldsymbol{\eta})$. BFGS progressively approximates the Hessian matrix by accumulating gradient information $\nabla g(\boldsymbol{\eta})$ at successive $\boldsymbol{\eta}$ points. Fortunately, $g(\boldsymbol{\eta})$ and $\nabla g(\boldsymbol{\eta})$ are both computable by solving the inner concave maximization on $\boldsymbol{\theta}$ (which in fact is equivalent to solving the primal minimization problem on $\mathbf{w}$). In particular, $g(\boldsymbol{\eta})$ is given by (10), and

$$\nabla g(\boldsymbol{\eta}) = \mathbf{c} + \frac{\log T}{2}\mathbf{e} - \frac{1}{2\beta}\sum_j \boldsymbol{\delta}_j^{*\cdot 2}$$

such that $\boldsymbol{\delta}_j^* = \sum_{\mathbf{b}_j} \#\mathbf{b}\left(\bar{\boldsymbol{\phi}}_{\mathbf{b}_j} - \boldsymbol{\theta}_{j\mathbf{b}}^*\right)$ for the optimal inner solution $\boldsymbol{\theta}^*$. Here, $\cdot^2$ denotes componentwise squaring.

## 7 Variable ordering

Our final step is to consider variable ordering as part of the optimization process. Once again, we will find that one can solve for the optimal ordering, while performing feature selection and parameter optimization simultaneously. Since no order is given, we first generate features for each variable $x_j$ assuming all other variables are potential parents. Then, as in the previous section, we introduce feature selection variables $\boldsymbol{\eta} = (...\eta_{jf}...)^\top$ and reduce the model complexity by minimizing the description length criterion.

As before, we begin by assuming the feature selection variables are $\{0,1\}$ valued. The issue now is that we need to add constraints to the $\boldsymbol{\eta}$ variables to ensure that a valid Bayesian network structure is obtained. For example, since activating a feature $\phi_{jf}$ for one variable means that the remaining variables in the pattern $f$ must be parents of $j$, no feature pattern $f$ can be activated for more than one variable $j \in f$. We can encode this constraint locally for each feature pattern by the constraints

$$\sum_{j\in f} \eta_{jf} \leq 1 \quad \text{for all } f \quad (11)$$

In fact, the local constraints are simple linear equalities that pose little additional burden on the optimization. Unfortunately, ensuring consistency locally within a feature pattern $f$ is easy, but ensuring consistency globally *between* feature patterns $f$ and $h$ is more difficult.

Our strategy for enforcing global consistency is to introduce a $\{0,1\}^{n\times n}$ matrix $S$ that encodes a total ordering on the variables. In particular, we let $S_{ij} = 1$ denote the case that $i$ precedes $j$ in the ordering, and $S_{ij} = 0$ denotes that $i$ follows $j$. For a matrix $S$ to encode a total ordering it has to be

antisymmetric: $\quad S_{ij} = 1 - S_{ji}$ for all $i \neq j \quad (12)$
transitive: $\quad S_{ij} + S_{jk} \leq S_{ik} + 1$ for all distinct $i, j, k$
reflexive: $\quad S_{ii} = 1$ for all $i$

(The diagonal of $S$ is not terribly important, but we set it 1 for convenience.) The feature selection variables can then be forced to respect a global ordering by the constraints

$$\eta_{jf} \leq S_{ij} \quad \text{for all} \quad f, i \in f, i \neq j \quad (13)$$

Thus, we obtain the result that for $\{0, 1\}$ valued variables $\boldsymbol{\eta}$ and $S$ we can enforce local and global consistency by *linear* constraints. This yields an obvious convex formulation for the entire relaxed problem.

$$\min_{\boldsymbol{\eta} \in [0,1]^F, S \in [0,1]^{n \times n}} g(\boldsymbol{\eta}) \text{ subject to (11), (12) and (13)}$$

One remaining problem with the formulation is that it requires a large, cubic number of constraints in (12) to encode the transitivity constraint. The cubic complexity can be reduced to a quadratic number of constraints by exploiting a few basic facts about relation matrices.

**Proposition 3** *Let $T$ be an upper triangular matrix with 1's above the main diagonal. A $\{0, 1\}$ valued matrix $S$ encodes a total ordering if and only if $S = I + U + (T - U)^\top$ for some $\{0, 1\}$ valued upper triangular $U$ above the diagonal such that $I + U + U^\top$ and $I + (T-U) + (T-U)^\top$ are both equivalence relations.*

**Proof** The idea is to show that transitivity is preserved in $\{0, 1\}$ matrices when one converts between an antisymmetric and symmetric relation. Let $T$ and $U$ be defined as above. Let $M = I + U + U^\top$, $N = I + (T - U) + (T - U)^\top$ and $S = I + U + (T - U)^\top$, and assume all values are in $\{0, 1\}$. Clearly $M$ and $N$ are symmetric and $S$ is antisymmetric.

We establish that for all $i, j, k$, such that $i \neq j$, $j \neq k$ and $i \neq k$, that $S_{ij} \wedge S_{jk} \Rightarrow S_{ik}$ if and only if $M_{ij} \wedge M_{jk} \Rightarrow M_{ik}$ and $N_{ij} \wedge N_{jk} \Rightarrow N_{ik}$. The argument is by cases over the six possible orderings of $i, j, k$.

**Case 1**: If $i < j < k$, then $S_{ij} = M_{ij}$, $S_{jk} = M_{jk}$, and $S_{ik} = M_{ik}$. Therefore $S_{ij} \wedge S_{jk} \Rightarrow S_{ik}$ iff $M_{ij} \wedge M_{jk} \Rightarrow M_{ik}$.

**Case 2**: If $i < k < j$, then $S_{jk} = N_{jk}$, $S_{ki} = N_{ki}$, and $S_{ji} = N_{ji}$. Therefore $S_{ij} \wedge S_{jk} \Rightarrow S_{ik}$ iff $\neg S_{ij} \vee \neg S_{jk} \vee S_{ik}$ iff $S_{ji} \vee \neg S_{jk} \vee \neg S_{ki}$ iff $N_{ji} \vee \neg N_{jk} \vee \neg N_{ki}$ iff $N_{jk} \wedge N_{ki} \Rightarrow N_{ji}$.

**Case 3**: If $k < i < j$, then $S_{ki} = M_{ki}$, $S_{ij} = M_{ij}$, and $S_{kj} = M_{kj}$. Therefore $S_{ij} \wedge S_{jk} \Rightarrow S_{ik}$ iff $\neg S_{ij} \vee \neg S_{jk} \vee S_{ik}$ iff $\neg S_{ij} \vee S_{kj} \vee \neg S_{ki}$ iff $\neg M_{ij} \vee M_{kj} \vee \neg M_{ki}$ iff $M_{ki} \wedge M_{ij} \Rightarrow M_{kj}$.

The remaining three cases are similar. ■

Thus, from the proposition, we can enforce transitivity by using a quadratic number of constraints by:

$$S = I + U + (T - U)^\top$$

$$\begin{aligned} I + U + U^\top &= DD^\top \\ E - U - U^\top &= CC^\top \\ D\mathbf{e} = \mathbf{e}, \quad C\mathbf{e} &= \mathbf{e} \end{aligned} \quad (14)$$

where $D$ and $C$ are further auxiliary square $\{0, 1\}$ matrices, and $E$ denotes the matrix of all 1s. Unfortunately, the two quadratic constraints are not convex, but they can be approximated by the semidefinite relaxations $I + U + U^\top \succeq DD^\top$, $E - U - U^\top \succeq CC^\top$. In our experiments below, we relaxed the $\{0, 1\}$ valued variables to $[0, 1]$ and used the semidefinite constraints. The implementation only requires a small modification to the BFGS strategy of the previous section, where the quasi-Newton step now needs to respect these constraints. We solved the convex constrained optimization problem by using a simple barrier method (Boyd & Vandenberghe, 2004), using a log barrier function for the linear inequality constraints (11) and (13), plus a log determinant barrier to enforce the semidefinite constraints in (14), thus ensuring a feasible search.

The result is the first comprehensive Bayesian network technique we are aware of that solves for an approximate variable ordering, feature set, and optimal weights in a joint, polynomial, convex optimization. Our results below show that this approach can produce competitive results.

One final issue to deal with is rounding a "soft" solution produced by the above convex optimization, to produce a variable ordering and a hard set of features to define a proper Bayesian network. We do not as yet have any approximation guarantees for any rounding approach we have developed so far. In our experiments below, we simply used a greedy rounding scheme that successively checks the largest non-integer $\eta$ variable, determines whether it is possible for it to be set to 1 without violating any consistency checks, and if so, rounds the variable greedily to 0 or 1 depending on which value yields the smallest value in the MDL objective (8) (keeping the current optimal Bayesian network parameters fixed). This is sufficient to yield reasonable results, although we would still like to investigate more sophisticated rounding approaches.

## 8 Experimental evaluation

We conducted a set of experiments on both synthetic and real data to evaluate our proposed algorithms and compare them to benchmark greedy heuristic search techniques. To measure performance of the different learning techniques, we measured the loglikelihood loss they achieve on held out test data after training. To isolate the effects of the different approximation stages, we conducted two sets of experiments: in the first set the variable ordering was held fixed, while in the second we used the relaxed ordering search of Section 7. In each case, for the greedy search algorithms, we considered both BDe and BIC scores.

For the fixed order experiments, we first considered three different artificial network structures, consisting of 5 variables and 4, 5, 6 edges respectively, each instantiated with random CPT entries. To set the regularization parameters for the convex relaxation technique we used an initial training sample and hold-out test sample, and then repeated the training and test procedures 10 times with the parameters fixed on independently generated data. We evaluated each learning technique by measuring the loglikelihood loss on an independent test set of size 1000 drawn from the same distribution as the training data. The results in the tables were averaged over the 10 repeats. We compared the results of the convex relaxation algorithm described in Section 6 to the K2 search algorithm of (Cooper & Herskovits, 1992), using both the BDe and BIC criteria as optimization objectives. All algorithms were given the correct variable ordering in these synthetic experiments.

Table 1: Synthetic experiments, comparing fixed order learning methods (given the correct variable order), with training sample size 50. Loglikelihood loss.

| Data Set | Convex | BIC | BDe |
|---|---|---|---|
| Synthetic 1 | 4.4753 | 4.5725 | 4.5264 |
| Synthetic 2 | 5.3385 | 5.4263 | 5.3854 |
| Synthetic 3 | 5.1641 | 5.2677 | 5.1788 |

Table 2: UCI data set experiments, comparing fixed order learning methods (given a random variable order), with training sample size 50. Loglikelihood loss.

| Data Set | Convex | BIC | BDe |
|---|---|---|---|
| breast | 5.0275 | 5.4698 | 5.2899 |
| cleve | 8.7238 | 8.9061 | 8.9984 |
| corral | 4.6109 | 4.6686 | 4.5084 |
| diabetes | 5.5130 | 5.6224 | 5.5942 |
| glass2 | 3.3533 | 3.5833 | 3.4047 |
| heart | 8.7866 | 8.8927 | 8.9570 |
| mofn | 7.6434 | 7.6734 | 7.8376 |
| pima | 5.2982 | 5.3460 | 5.3635 |

Table 3: Synthetic experiments, comparing methods that learn both structure and order, with training sample size 50. Loglikelihood loss.

| Data Set | Convex | BIC | BDe |
|---|---|---|---|
| Synthetic 1 | 4.4887 | 4.5731 | 4.5072 |
| Synthetic 2 | 5.3413 | 5.4265 | 5.3489 |
| Synthetic 3 | 5.1581 | 5.2692 | 5.2105 |

Table 4: UCI data set experiments, comparing methods that learn both structure and order, with training sample size 50. Loglikelihood loss.

| Data Set | Convex | BIC | BDe |
|---|---|---|---|
| breast | 5.2643 | 5.2289 | 5.2491 |
| cleve | 8.5974 | 8.8418 | 9.1324 |
| corral | 4.7056 | 4.5577 | 4.5360 |
| diabetes | 5.5823 | 5.6098 | 5.6210 |
| glass2 | 3.4870 | 3.5803 | 3.3950 |
| heart | 8.5889 | 8.8684 | 9.0956 |
| mofn | 7.5659 | 7.7125 | 8.0055 |
| pima | 5.3823 | 5.3369 | 5.3885 |

Table 1 shows the results obtained by the convex relaxation technique versus the greedy search algorithms on a training sample of size 50 drawn from the synthetic Bayesian network models. Here we can see that the convex technique outperforms the greedy heuristic search procedures, using both the BDe and BIC scores. However, the run time for the convex relaxation procedure (including rounding) was 10, 25 and 30 seconds respectively, while the K2 algorithm on required 0.05 seconds on these problems.

Next, we conducted an experiment on real data. Here we used several UCI data sets: corral, breast, cleve, diabetes, glass2, heart, pima and mofn. For each dataset, we ran the learning algorithms 10 times on different random training/test partitions, using the training set for Bayesian network learning and test set for performance evaluation. Each algorithm was run with the same fixed variable order, where in this case the order was just chosen randomly. For the convex relaxation technique, we used one preliminary training/test split to set the regularization parameters. Table 2 shows the average loglikelihood loss obtained on the test partitions, training on a disjoint subset of size 50 examples, for each of the learning methods. Here we see that the convex approach holds an advantage over the greedy search techniques, for both BDe and BIC scores. Once again, however, the run times of the convex relaxation approach are greater than the K2 algorithm, requiring from 10-100 times more time to produce the final results.

Although these results are preliminary, they suggest that the ability to avoid local minima in a discrete structure search can lead to good solutions. On the other hand, the major disadvantages for the approach we have presented so far is that it runs slower than heuristic greedy search and requires regularization parameters to be set, where as the BDe and BIC scores are parameter free. Beyond improving run time, one significant direction for future research is to reduce the reliance on regularization parameters.

Next, we repeated the previous experiments using the combined structure and order optimization algorithm of Section 7. Here we compared to greedy heuristic search using the standard edge addition, deletion and reversal operators, again considering both the BDe and BIC optimization criteria. Each greedy search was started from an empty network and restarted 4 times when reaching a local minimum, by randomly adding and deleting edges. However, other than not imposing a variable ordering, the algorithms were run exactly as described above for the fixed order case.

Table 3 shows the results obtained by the convex relaxation technique versus the greedy search algorithms on the synthetic problems. Here we see a modest advantage for the convex over greedy search methods. However, once again, the convex relaxation procedure runs about 100 times slower. Interestingly, the solution quality is close to the fixed order case, which only benefited slightly from having the correct variable ordering.

Table 4 shows the results obtained by the convex relaxation technique versus the greedy search algorithms on the UCI data sets. Here the quality of the outcome is mixed. The convex relaxation procedure obtains the best solution quality on 4 out of 8 data sets, while the greedy heuristic search using BDe obtains the best results on 2 out of 8, and BIC obtains the best results on 2 out of 8. More interestingly, comparing these results to the fixed order technique, which just uses a random variable ordering, shows that the fixed order approach (with convex relaxation) still obtains the best results on 4 out of 8 data sets. This outcome seems to suggest that the relaxed ordering constraints imposed in Section 7 might not be sufficiently tight to ensure a good solution. Improving the quality of the relaxed ordering constraints remains an important question for future research.

## 9 Conclusion

We have presented what we feel is a promising new perspective on learning Bayesian network structure from data. The technique simultaneously searches for variable order, parameter settings, and features in a joint convex optimization. We feel that this approach might open the way to a new class of algorithms for learning Bayesian networks that ultimately might lead to guaranteed approximation quality. Beyond approximation guarantees and algorithmic improvements, other significant directions for future research include considering the problem of learning in the presence of missing data or latent variables (Elidan & Friedman, 2005), and attempting to extend the current analysis to Bayesian scores (Cooper & Herskovits, 1992; Heckerman et al., 1995).